\documentclass[letterpaper, 10 pt, conference]{ieeeconf}  

\IEEEoverridecommandlockouts                              

\usepackage{cite}
\usepackage{amsmath,amssymb,amsfonts}
\usepackage{algorithmic}
\usepackage{float}
\usepackage{graphicx}
\usepackage{multicol}
\usepackage{subcaption}
\usepackage{textcomp}
\usepackage{bm}
\usepackage[ruled,vlined]{algorithm2e}
\usepackage{hyperref}

\def\Vec#1{\!\!\hbox{$#1$\kern-0.38em\lower0.85em\hbox{$\vec{}\,$}}\,}%
\newcommand{\bbm}{\begin{bmatrix}}
	\newcommand{\ebm}{\end{bmatrix}}
\DeclareMathAlphabet{\mbf}{OT1}{ptm}{b}{n}

\newcommand{\ignore}[1]{}

\newcommand{\bma}[1]{\left[\begin{array}{#1}}
\newcommand{\ema}{\end{array}\right]}

\DeclareMathAlphabet{\mbf}{OT1}{ptm}{b}{n}

\def\fdotb{{\raisebox{-0.6ex}{ \kern0.2ex\raisebox{0.8ex}{\tiny $\hspace*{-1ex}\circ$}}}}
\def\fddotb{{\raisebox{-0.6ex}{ \kern0.2ex\raisebox{0.8ex}{\tiny $\hspace*{-1ex}\circ\circ$}}}}

\newcommand{\utimes}{ {\raisebox{-0.6ex}{ \kern-1.0ex\raisebox{0.6ex}{ \small $\mathsf{v}$}}} } %
\newcommand{\beq}{\begin{equation}}
\newcommand{\eeq}{\end{equation}}
\newcommand{\bdis}{\begin{displaymath}}
\newcommand{\edis}{\end{displaymath}}
\newcommand{\beqarray}{\begin{eqnarray}}
\newcommand{\eeqarray}{\end{eqnarray}}
\newcommand{\beqarraynn}{\begin{eqnarray*}}
\newcommand{\eeqarraynn}{\end{eqnarray*}}

\title{\LARGE \bf
Uncertainty-Constrained Differential Dynamic Programming in Belief Space for Vision Based Robots
}

\author{Shatil Rahman*, Steven L. Waslander*
\thanks{* This research was Supported by the Natural Sciences and Engineering
Research Council of Canada (NSERC) and NSERC Canadian Robotics Network (NCRN).
S. Rahman and S.L. Waslander are with the University of Toronto Institute
for Aerospace Studies and Robotics Institute.
        {\tt\small \{shatil.rahman@mail.utoronto.ca\}, \{stevenw@utias.utoronto.ca\}}}%
}

\begin{document}

\maketitle
\thispagestyle{empty}
\pagestyle{empty}

\begin{abstract}
	Most mobile robots follow a modular sense-plan-act system architecture that can lead to poor performance or even catastrophic failure for visual inertial navigation systems due to trajectories devoid of feature matches. Planning in \textit{belief space} provides a unified approach to tightly couple the perception, planning and control modules, leading to trajectories that are robust to noisy measurements and disturbances. However, existing methods handle uncertainties as costs that require manual tuning for varying environments and hardware. We therefore propose a novel trajectory optimization formulation that incorporates inequality constraints on uncertainty and a novel Augmented Lagrangian based stochastic differential dynamic programming method in belief space. Furthermore, we develop a probabilistic visibility model that accounts for discontinuities due to feature visibility limits. Our simulation tests demonstrate that our method can handle inequality constraints in different environments, for holonomic and nonholonomic motion models with no manual tuning of uncertainty costs involved. We also show the improved optimization performance in belief space due to our visibility model.

\end{abstract}


\section{Introduction}
Motion planning under uncertainty provides a unified approach to tightly couple the perception and action modules, leading to trajectories and policies that are robust to noisy measurements and disturbances. Vision based robots in particular can suffer from localization failure due to rapid perspective shifts or trajectories through regions devoid of features. In this paper we leverage planning in \textit{belief space}, a space of probability distributions. Similar to previous work\cite{van2012motion}, we approximate the belief as Gaussian distributions and use the Extended Kalman Filter equations to define belief dynamics. Thus the resulting continuous Partially Observable Markov Decision Process (POMDP) is formulated as a stochastic optimal control problem that minimizes a cost in expectation subject to belief dynamics. This is solved via a differential dynamic programming (DDP) approach to obtain a locally optimal trajectory and locally linear optimal policy. 
	
Due to the explicit inclusion of the covariance matrix elements in the belief state, the cost function can impose scalar information costs on uncertainty, such as the trace or determinant of the covariance matrix. However, the choice of information cost, as well as its weight relative to other costs leads to unpredictable optimal trajectories which necessitates careful, application-specific tuning. We propose that the true benefit of planning in belief spaces is not to impose costs on uncertainty, but to enforce \textit{constraints} on the uncertainty, which is explicitly a part of the belief state. However, previous belief space formulations based on dynamic programming cannot handle such inequality constraints, and directly using optimization solvers would only lead to open loop controls which are not robust to noise or disturbances unless used as a part of a Model Predictive Control (MPC) framework, which may not be computationally feasible, especially with regards to the large state space size due to inclusion of the covariance. Furthermore, trajectory optimization methods in belief space require smooth motion and measurement models, which are not applicable for feature based visual localization systems. This paper proposes a novel constrained stochastic DDP approach in belief space that
\begin{itemize}
	    \item formulates motion planning under uncertainty as an uncertainty constrained stochastic optimal control problem,
	    \item uses an augmented Lagrangian based Differential Dynamic Programming approach to handle linear constraints on the elements of the covariance matrix and drive the solver towards constraint satisfaction and
	    \item leverages Expectation-Maximization with binary measurement acquisition variables to probabilistically account for the effect of camera field of view (FoV) limits and perspective shifts in a smooth manner amenable to gradient-based optimization.
\end{itemize}
Our approach can incorporate constraints on uncertainty, which are intuitive and have quantifiable consequences on safety, such as desired upper limits on $3\sigma$ covariance bounds, while retaining the advantages of DDP methods such as providing a closed loop linear feedback policy around the resulting optimal trajectory and computational efficiency that scales linearly with horizon length. It also provides an explicit \textit{a priori} estimate of the maximum constraint violation for infeasible problems to enable `go or no go' decision making. The ability to incorporate camera measurements allows our method to be used in many cluttered indoor environments such as factories and warehouses with only visual navigation. Our simulation results for holonomic and non-holonomic vehicles equipped with stereo cameras demonstrate that our method is able to satisfy uncertainty constraints with no manual tuning in varying environments and that our smooth visibility modelling leads to a lower local optimal cost.

\section{Related Work}\label{sec:related}
In its most general form, the planning under uncertainty problem is a Partially Observable Markov Decision Process (POMDP)~\cite{hauskrecht2000value}. A vast literature exists on general purpose POMDP solvers~\cite{pineau2003point,chaudhari2013sampling,somani2013despot}. These methods discretize the state space, pick subsets of their corresponding beliefs, simulate forward in time to create search trees and carry out value iteration to obtain optimal policies. Despite recent solvers that can handle thousands of discrete states~\cite{shani2013survey}, the curse of dimensionality renders these methods unsuitable for large continuous state-action-observation spaces in robotics such as for UAVs with cameras and IMUs.
    
To apply belief space planning in robotics problems, the beliefs are commonly approximated as Gaussians. These methods can be split into sampling based and optimization based planners. Seminal work in the former category include Belief Roadmaps~\cite{prentice2009belief} which builds a roadmap in state space, projects it into belief space given a starting belief and finds optimal paths in the resulting graph. However the resulting path is only optimal relative to the initial roadmap and cannot be re-queried for a different starting or final belief. Rapidly exploring random belief trees~\cite{bry2011rapidly} use heuristics to prune a tree projected in belief space to find a globally optimal path, however it assumes fixed controller gains along edges. To facilitate replanning and multiple queries, Feedback Based Information Roadmaps ~\cite{agha2011firm} and extensions thereof~\cite{agha2015simultaneous} first design stationary and time varying belief controllers offline around PRM nodes and edges, respectively, to discretize the continuous problem, to be solved online efficiently when needing to replan. Again, resulting policies are only resolution optimal and still assume fixed controllers. Furthermore, the initial sampling is always done in state space, irrespective of the utility of the points from an information perspective. 
    
In contrast to sampling based methods which look for globally (resolution) optimal paths, optimization methods seek locally optimal belief trajectories and controls around a given trajectory in state space. Early works~\cite{platt2010belief,erez2012scalable} in this category maintain an augmented state consisting of the belief mean and covariance and formulate \textit{belief dynamics}, which are then used in a control approach. However, they assume that future observations will be exactly the maximum likelihood observation, making the belief dynamics deterministic. Van der Berg \textit{et al.}~\cite{van2012motion} maintain the stochasticity of the belief dynamics and solve using iterative LQG control. These methods require smooth motion and measurement models, and so cannot readily be used for vision based systems. Furthermore, they cannot handle inequality constraints on uncertainty states. More recently, online MPC-based approaches~\cite{indelman2015planning,chaves2015risk} have been prevalent. These methods first compute beliefs over entire trajectories and then optimize over them. However, these methods are computationally expensive due to the inclusion of a large augmented belief state over a trajectory as well as the need to compute derivatives using finite differences. Finally, the method does not consider obstacle avoidance or camera field of view limitations.


\section{Problem Definition} \label{sec:def}
 Let $\mbf{x} \in \mathbb{R}^{n}$ be the true state of the robot, which can take inputs $\mbf{u} \in \mathbb{R}^{m}$ and receive measurements of the world $\mbf{z} \in \mathbb{R}^{p}$. At each timestep $k, \{k = 0,\ldots,K-1\}$, the robot is subject to stochastic Markov dynamics and measurements according to
 \begin{align}
   \mbf{x}_{k+1} &=  \mbf{f}_k(\mbf{x}_{k},\mbf{u}_{k},\mbf{v}_k),\label{eq:dyn}\\
   \mbf{z}_{k}  &= \mbf{y}_k(\mbf{x}_{k},\mbf{n}_k),
\label{eq:meas}
\end{align}
where $\mbf{v}_k \sim \mathcal{N}(\mbf{0},\mbf{Q}_k)$ and  $\mbf{n}_k \sim \mathcal{N}(\mbf{0},\mbf{R}_k)$ are independent noise variables. In this work, the world is static and consists of free space, $X_{free}$, and space occupied by obstacles, $X_{obs}$. The exteroceptive measurements in \eqref{eq:meas} are assumed to be from cameras, although our method can be generalized to any view or region limited observations. More specifically, let $\mbf{p}_l, l = 1,\ldots,L$, be the position of the $l$\textsuperscript{th} visual feature, expressed in the world frame. Then the pixel measurements of $\mbf{p}_l$ at time $k$ are given by
    \begin{equation}\label{eq:cam_meas}
        \mbf{z}_{l,k} = \mbf{h}(\mbf{T}_{ki}(\mbf{x}_k)\mbf{p}_l) + \mbf{n}_{l,k},
    \end{equation}
where $\mbf{T}_{ki}: \mathbb{R}^3 \to SE(3)$ returns the transformation matrix that transforms points in the world frame to the robot camera frame at time $k$, and $\mbf{h}: SE(3)\to \mathbb{R}^2$ projects points to pixel coordinates in the camera image plane, and is specific to the type of camera (monocular, stereo etc). Although it is restrictive to assume that the true feature locations, as well as $X_{free}$ and  $X_{obs}$ are known, this assumption is consistent with a pre-mapped operating model, such as occurs in industrial inspection or transport routes. We also restrict feature matches to when the robot is within a cone of when the feature was originally mapped.
    
The Gaussian belief of the robot at time $k$ is denoted by $\mbf{b}_k$, parameterized as $\mbf{b}_k = [\hat{\mbf{x}}_k,\mathrm{vec}(\bm{\Sigma}_k)]$, where $\hat{\mbf{x}}_k$ and $\bm{\Sigma}_k$ are the mean and covariances respectively, and the $\mathrm{vec}(\cdot)$ stacks the columns of the $n \times n$ covariance matrix, accounting for symmetry by not duplicating off-diagonal entries, following the notation of Magnus and Neudecker~\cite{magnus2019matrix}. Central to this formulation are the stochastic \textit{belief dynamics},
    \begin{equation}\label{eq:bel_dyn}
        \mbf{b}_{k+1} = \mbf{g}(\mbf{b}_{k},\mbf{u}_{k}) + \mbf{w}(\mbf{b}_k,\mbf{u}_k)\bm{\xi}_k,
    \end{equation}
where $\mbf{g}(\cdot)$ are the deterministic dynamics and $\mbf{w}(\cdot)\bm{\xi}$ is state and control dependant noise, whose properties will be described in Section \ref{sec:belief_dyn}. These dynamics are estimator specific and depend on \eqref{eq:dyn} and \eqref{eq:meas}.
Thus, the motion planning under uncertainty problem can be formulated as
\begin{equation}\label{big_opt}
    \begin{aligned}
        &  \min\limits_{\mbf{u}_{0:K-1}}
        & & J(\mbf{b}_{0:K},\mbf{u}_{0:K-1}) = \mathbb{E}\{c_K(\mbf{b}_K) + \sum_{k=0}^{K-1}c_k(\mbf{b}_k,\mbf{u}_k)\}. \\
        &   \text{s.t.}
        &  &\mbf{b}_{k+1} = \mbf{g}(\mbf{b}_{k},\mbf{u}_{k}) + \mbf{w}(\mbf{b}_k,\mbf{u}_k)\bm{\xi}_k,\\
        & &  &\bm{\psi}_{k}(\mbf{b}_{k}) < \mbf{0}
    \end{aligned}
\end{equation}
where $c_K(\mbf{b}_K)$ and $c_k(\mbf{b}_k,\mbf{u}_k)$ are terminal and stage costs and $\bm{\psi}_{k}(\mbf{b}_{k},\mbf{u}_k)$ is a column matrix of $J$ constraints at timestep $k$. Although our method can handle general nonlinear inequality constraints, due to the inclusion of the covariance in the belief state it is easy to see that desired maximum uncertainty constraints appear as simple state bound constraints.


\section{Probabilistic Visibility Based Belief Dynamics}\label{sec:prob_dyn}

\subsection{Belief Dynamics}\label{sec:belief_dyn}
We shall now present the belief dynamics for a system with motion and measurement models of the form of \eqref{eq:dyn} and \eqref{eq:meas}. The derivation is similar to that by Van der Berg \textit{et al.}~\cite{van2012motion}, however here we explicitly derive the deterministic and stochastic components of the dynamics due to estimation, motion and measurement uncertainty. First, we perform a first order Taylor series expansion of the motion and measurement models about $\hat{\mbf{x}}_k$ to obtain
\begin{equation}\label{eq:lin_mot}
    \mathbf{x}_{k+1} \approx \check{\mathbf{x}}_{k+1}+\mathbf{F}_{k}\left(\mathbf{x}_{k}-\hat{\mathbf{x}}_{k}\right)+\mathbf{v}_{k}^{\prime},
\end{equation}
\begin{equation}\label{eq:lin_meas}
    \mathbf{z}_{k+1} \approx \check{\mathbf{z}}_{k+1}+\mathbf{H}_{k}\left(\mathbf{x}_{k+1}-\check{\mathbf{x}}_{k+1}\right)+\mathbf{n}_{k}^{\prime},
\end{equation}
where $\check{\mathbf{x}}_{k+1},\check{\mathbf{z}}_{k+1}$ are the predicted mean and measurement. Substituting $  \mbf{e}_{k+1} = \mathbf{x}_{k+1} -  \check{\mathbf{x}}_{k+1}$ from \eqref{eq:lin_mot} into \eqref{eq:lin_meas} gives us
\begin{equation}\label{eq:almost_z}
    \mathbf{z}_{k+1} \approx \check{\mathbf{z}}_{k+1}+\mathbf{H}_{k}\mathbf{F}_{k}\mbf{e}_k+\mathbf{H}_{k}\mathbf{v}_{k}^{\prime}+\mathbf{n}_{k}^{\prime}.
\end{equation}
 
 The equations of the Extended Kalman Filter can now be used to formulate the belief dynamics. The update equation of the EKF for the mean is
 $$
    \hat{\mbf{x}}_{k+1} = \mbf{f}(\hat{\mbf{x}}_{k},\mbf{u}_{k},\mbf{0}) + \mbf{K}_{k}(\mathbf{z}_{k+1} - \check{\mathbf{z}}_{k+1}),
 $$
 where $\mbf{K}_k$ is the standard Kalman gain. 
 Substituting the innovation term $\mathbf{z}_{k+1} - \check{\mathbf{z}}_{k+1}$ from \eqref{eq:almost_z} into the update equation gives us 
 \begin{equation}\label{eq:mean_update}
     \hat{\mbf{x}}_{k+1} = \mbf{f}(\hat{\mbf{x}}_{k},\mbf{u}_{k},\mbf{0}) + \mbf{m}(\mbf{b}_{k},\mbf{u}_{k}) \bm{\xi}_k
 \end{equation}
 where
 $$
    \mbf{m}(\mbf{b}_{k},\mbf{u}_{k}) = [\mbf{K}_k \mbf{H}_{k}\mbf{F}_{k} \quad \mbf{K}_k \mbf{H}_{k} \quad \mbf{K}_k]
 $$
 and $\bm{\xi}^{T}_k = [\mbf{e}^T_k,\mathbf{v}_{k}^{\prime T}, \mathbf{n}_{k}^{\prime T}]$. 
 
 The covariance update of the EKF is simply
 \begin{equation}\label{eq:cov_update}
     \bm{\Sigma}_{k+1} = \left(\mathbf{1}-\mathbf{K}_{k} \mathbf{H}_{k}\right)\left(\mathbf{F}_{k} \bm{\Sigma}_{k} \mathbf{F}_{k}^{T}+\mathbf{Q}_{k}\right).
 \end{equation}
Equations \eqref{eq:mean_update} and \eqref{eq:cov_update} can be put together to form \eqref{eq:bel_dyn}
where
 $$
    \mbf{g}(\mbf{b}_{k},\mbf{u}_{k},\mbf{w}_k) = \bma{c}
                \mbf{f}(\hat{\mbf{x}}_{k},\mbf{u}_{k},\mbf{0})\\
                \mathrm{vec}\left(\left(\mathbf{1}-\mathbf{K}_{k} \mathbf{H}_{k}\right)\left(\mathbf{F}_{k} \bm{\Sigma}_{k} \mathbf{F}_{k}^{T}+\mathbf{Q}_{k}\right)\right)
                \ema,
 $$
and
$$
\mbf{w}(\mbf{b}_k,\mbf{u}_k) = \bma{c}
                        \mbf{m}(\mbf{b}_k,\mbf{u}_k)\\
                        \mbf{0}\ema.
$$
\subsection{Probabilistic Visibility Modelling}\label{sec:prob_mod}
Belief space planning via gradient-based trajectory optimization requires smooth belief dynamics. However, camera measurement models present discontinuities as the uncertainty can vary rapidly as features come in and out of view~\cite{van2012motion}. Numerical approximations of gradients across discontinuous cost boundaries result in artificially large gradients that hinder optimization. 

In this paper, we propose a modification to the camera model in \eqref{eq:cam_meas} that results in belief dynamics that are smooth. We follow the approach of Indelman \textit{et al.}~\cite{indelman2015planning} in modelling measurement acquisition by introducing Bernoulli random variables $\bm{\Gamma}_k = [\gamma_{1,k},\ldots,\gamma_{L,k}]$, where $p(\gamma_{l,k}=1)$ represents the probability of the $l$\textsuperscript{th} feature being matched by the perception stack at time $k$. The mean of the posterior belief $\hat{\mbf{x}}_{k+1}$ can be obtained from maximum a posteriori estimation approach by maximizing $p(\mbf{x}_{k+1},\bm{\Gamma}_{k+1}|\mbf{z}_{k+1},\mbf{x}_{k},\mbf{u}_k)$, where $\bm{\Gamma}_k$ is now present as a hidden variable. To address the hidden variables, $\bm{\Gamma}_k$, we leverage \textit{expectation-maximization}~\cite{minka1998expectation} (EM) as follows  
\begin{multline}
    \hat{\mbf{x}}_{k+1} = \underset{\mbf{x}_{k+1}}{\mathrm{max}} \quad  \underset{\bm{\Gamma}_{k+1}|\bar{\mbf{x}}_{k+1}}{\mathbb{E}}[\ln( p(\mbf{z}_{k+1},\bm{\Gamma}_{k+1}|\mbf{x}_{k+1}) \\
    \times p(\mbf{x}_{k+1}|\mbf{x}_{k},\mbf{u}_k))]
\end{multline}\begin{multline}
    = \underset{\mbf{x}_{k+1}}{\mathrm{min}} -\sum_{l=1}^{L}p(\gamma_{l,k+1}=1|\bar{\mbf{x}}_{k+1})\ln p(\mbf{z}_{k+1}|\mbf{x}_{k+1}) \\
     -\ln p(\mbf{x}_{k+1}|\mbf{x}_{k},\mbf{u}_k),\label{expect}
\end{multline}
where the expectation is taken with respect to the hidden variables, given an initial guess $\check{\mbf{x}}_{k+1}$ which can be the predicted mean. Notice that in \eqref{expect}, upon taking the expectation with respect to hidden variable $\gamma_{l,k}$, the term with $\gamma_{l,k} = 0$ vanishes. Normally in EM, this procedure is iterated with the new estimate $\hat{\mbf{x}}_{k+1}$ being the guess for the expectation, but here we restrict our formulation to a single EM step. Considering the EKF can also be thought of as one iteration of an MAP estimation approach which shows good results in practice, this approximation seems reasonable, and may be explored further in future work. 

From the first term in the summation in \eqref{expect} we can see that simply scaling the measurement likelihood by the probability of a feature being detected is equivalent to the expectation maximization procedure, which can be done by scaling the measurement covariance by $\tilde{\mbf{R}}_{l}^{-1} = p_l \mbf{R}_{l}^{-1}$.  Intuitively, this is desirable, as a measurement with $p(\gamma = 0)$ corresponds to a measurement with zero information, and will thus not contribute to the state estimate. 

Finally, in this work, we propose the following model for $p(\gamma_{l,k}=1|\mbf{x}_{k})$. Let $\hat{\mbf{e}} = [0,0,1]^{T}$ be the camera z-axis and $\hat{\mbf{e}}^{l}_i$ be the surface normal unit vector of the $l$\textsuperscript{th} feature expressed in the world frame.  Furthermore let $\mbf{p}_{c}^{lc}$ be the position of the $l$\textsuperscript{th} feature relative to the camera expressed in the camera frame and $\mbf{p}_{i}^{kl}$ be the vector from the $l$\textsuperscript{th} feature to the $k$\textsuperscript{th} robot position on in the world frame. Then
\beq \label{vis}
p(\gamma_{l,k}=1|\mbf{x}_{k}) =\left\{\begin{array}{ll}{p_1(\alpha_{l,k})p_2(\beta_{l,k})} & {|\alpha_{l,k}|<\alpha_{\max }} \\  & {\mathrm{and}\;|\beta_{l,k}|<\beta_{\max }} \\ {0} & {\text { else }}\end{array}\right.
\eeq
with 
\begin{align*}
p_1(\alpha_{l,k}) &= \frac{1}{2}\left(\cos \frac{\alpha_{l,k}}{\alpha_{\max }} \pi+1\right),\\
p_2(\beta_{l,k}) &= \frac{1}{2}\left(\cos \frac{\beta_{l,k}}{\beta_{\max }} \pi+1\right),\\
\alpha_{l,k}&=\cos ^{-1}\left(\frac{\mathbf{p}_{c}^{lc \top} \hat{\mathbf{e}}}{\left\|\mathbf{p}_{c}^{lc}\right\|_{2}}\right) \quad \in[0, \pi],\\
\beta_{l,k}&=\cos ^{-1}\left(\frac{\mathbf{p}_{i}^{kl \top} \hat{\mbf{e}}^{l}_i}{\left\|\mathbf{p}_{i}^{kl}\right\|_{2}}\right) \quad \in[0, \pi].\\
\end{align*}
Intuitively, $\alpha$ is the angle between the camera z-axis and the $l$\textsuperscript{th} feature, and $\beta$ is the angle between the robot position and the normal direction associated with the $l$\textsuperscript{th} feature, which was obtained during mapping, that is, a measure of perspective shift. $\alpha_{max}$ approximates the field of view of the camera and $\beta_{max}$ approximates the largest perspective shift possible before feature matching fails. The probability of visibility and thus the information entry for $\mbf{z}_{l,k}$ smoothly drops off to zero at the boundary, which ensures that the belief dynamics remain continuous. Finally, we note that this will result in a conservative covariance prediction compared to an actual estimator, as information from measurements are down-weighted near the boundary of the field of view.

\section{Constrained Stochastic Differential Dynamic Programming in Belief Space}\label{sec:meth}

To solve the belief space planning problem, posed as the constrained stochastic optimal control problem in \eqref{big_opt}, we opt for a differential dynamic programming (DDP) approach similar to Van der Berg \textit{et al}\cite{van2012motion}. However, to deal with the state constraints, we adopt an Augmented Lagrangian~\cite{lantoine2012hybrid}\cite{aoyama2020constrained} constrained DDP approach, which we present in detail in this section.

We proceed by applying a belief space variant of the iterative Linear Quadratic Gaussian (iLQG) method \cite{todorov2005generalized}, a type of DDP algorithm for unconstrained optimal control problems. The main idea is to reduce the optimization problem over a horizon $\{1:K\}$ to a series of optimizations at each timestep by leveraging \textit{Bellman's principle of optimality} and to solve these optimizations iteratively backward through time for a feedback policy $\mbf{u}_{k} = \bm{\pi}_{k}(\mbf{b}_{k}), k = 1,\ldots,K-1$.

Augmented Lagrangian approaches solve an optimization problem subject to nonlinear equality and inequality constraints by solving a sequence of unconstrained problems. As the name suggests, this is done by augmenting the cost function with a \textit{penalty function}. Specifically, we wish to solve

\begin{equation}\label{eq:aug_prob}
        \begin{aligned}
            &  \underset{\mbf{u}_{0:K-1}}{\text{min}}
            & & J(\mbf{b}_{1:K},\mbf{u}_{0:K-1}) + \sum_{k=0}^{K} \mathcal{P}\left(\bm{\lambda}_{k}, \bm{\mu}_{k}, \bm{\psi}_{k}\left(\mbf{b}_{k}, \mbf{u}_{k}\right)\right)\\
            &   \text{s.t.}
            &  &\mbf{b}_{k+1} = \mbf{g}(\mbf{b}_{k},\mbf{u}_{k}) + \mbf{w}(\mbf{b}_k,\mbf{u}_k)\bm{\xi}_k, \\
        \end{aligned}
\end{equation}
$\bm{\lambda}_k \in \mathbb{R}^{J}$ are estimates of the Lagrange multipliers of the original constrained problem and $\bm{\mu}_k \in \mathbb{R}^{J}$ are penalty parameters. The role of the penalty function is to impose a cost for violating the constraints $\bm{\psi}_{k}$ and the weight of this cost depends on the Lagrange multipliers $\bm{\lambda}_k$ and the penalty parameters $\bm{\mu}_k$. The penalty function we opt for in this paper is
\begin{equation}\label{eq:penalty_func}
    \mathcal{P}\left(\bm{\lambda}_{k}, \bm{\mu}_{k}, \bm{\psi}_{k}\left(\mbf{b}_{k}, \mbf{u}_{k}\right)\right) = \sum^{J}_{j=1}\frac{\left(\lambda_{j, k}\right)^{2}}{\mu_{j, k}} \phi\left(\frac{\mu_{j, k}}{\lambda_{j, k}} {\psi}_{j,k}\left(\mbf{b}_{k}, \mbf{u}_{k}\right)\right)
\end{equation}
where ${\psi}_{j,k}$ is the $j$\textsuperscript{th} row of the constraint function $\bm{\psi}_{k}$ and
$$
\phi(t):=\left\{\begin{array}{ll}
\frac{1}{2} t^{2}+t, &  t \geq-\frac{1}{2} \\
-\frac{1}{4} \log (-2 t)-\frac{3}{8}, & \text { otherwise }
\end{array}\right.
$$
This is a smooth approximation of the popular Powell-Hestenes-Rockafellar penalty function\cite{nocedal2006numerical}\cite{aoyama2020constrained} which is quadratic when the violation is large and decays quickly to zero when the violation is small or there is no violation.

 It can be shown that under mild assumptions on $\mathcal{P}(\cdot)$, a solution of \eqref{eq:aug_prob} is equivalent to a solution to the original constrained problem\cite{nocedal2006numerical}. Our Augmented Lagrangian based iLQG method (iLQG-AL), works in a similar manner and consists of an ``inner" and ``outer" loop, which we describe in detail in Sections \ref{sec:inner} and \ref{sec:outer}. 

The method is initialized with an initial belief and control trajectory $\{\bar{\mbf{b}}_{1:K}, \bar{\mbf{u}}_{1:K-1}\}$ which can be obtained by using a state space planner to generate open loop controls and applying \eqref{eq:bel_dyn} with zero noise to form the nominal belief trajectory. Furthermore, $\bm{\lambda}_{1:K}$ and  $\bm{\mu}_{1:K}$ are initialized to arbitrary positive values.

\subsection{Inner Loop: Solving the unconstrained problem}\label{sec:inner}
Here we derive the equations required to iteratively solve \eqref{eq:aug_prob}. For brevity, let
\begin{equation}
     \tilde{c}_k(\mbf{b}_k) =  c_k(\mbf{b}_k) + \mathcal{P}\left(\bm{\lambda}_{k}, \bm{\mu}_{k}, \bm{\psi}_{k}\left(\mbf{b}_{k},\mbf{u}_k\right)\right).
\end{equation}
 Central to the method is the \textit{value function} which represents the minimum cost-to-go at each belief state and time, defined as 
\begin{align}
    V_K(\mbf{b}_K) &= \tilde{c}_K(\mbf{b}_K),\\
    V_k(\mbf{b}_k) &=  \underset{\mbf{u}_{k}}{\text{min}} \quad \left\{\tilde{c}_k(\mbf{b}_k,\mbf{u_k}) + \mathbb{E}[V_{k+1}(\mbf{b}_{k+1})]\right\}.\label{value_func}
\end{align}
DDP methods find local solutions to \eqref{value_func} at each timestep by performing Taylor expansions on the terms in the right hand side about $\{\bar{\mbf{b}}_{1:K}, \bar{\mbf{u}}_{1:K-1}\}$. Specifically, let
\begin{equation}\label{eq:q_func}
    Q_k(\mbf{b}_k,\mbf{u}_k) = \tilde{c}_k(\mbf{b}_k,\mbf{u_k}) + \mathbb{E}[V_{k+1}(\mbf{g}(\mbf{b}_{k},\mbf{u}_{k}) + \mbf{w}(\mbf{b}_k,\mbf{u}_k)\bm{\xi})]
\end{equation}
denote the argument of the minimization in \eqref{value_func}, where we have substituted in \eqref{eq:bel_dyn}. For ease of notation, let $S_{\mbf{x}}$ and $S_{\mbf{x}\mbf{x}}$ denote the gradient and Hessian matrices respectively for any smooth scalar function $S(\mbf{x})$ and let $\mbf{y}_{\mbf{x}}$ denote the Jacobian matrix for any smooth vector function $\mbf{y}(\mbf{x})$, all evaluated at $\mbf{x} = \bar{\mbf{x}}$.

We begin at the last timestep by making a quadratic approximation of $V_K$,
\begin{align}
    V_K &\approx \bar{V}_K + V_{\mbf{b}_K}^{T}\delta\mbf{b}_K + \frac{1}{2}\delta\mbf{b}_K^{T}V_{\mbf{bb}_K}\delta\mbf{b}_K \\
    &= \tilde{c}_K(\mbf{\bar{b}}_K) + \tilde{c}_{\mbf{b}_K}^{T}\delta\mbf{b}_K + \frac{1}{2}\delta\mbf{b}_K^{T}\tilde{c}_{\mbf{bb}_K}\delta\mbf{b}_K.\label{eq:v_quad}
\end{align}
\begin{figure*}[ht]
    \centering
        \begin{subfigure}{0.47\textwidth}
            \includegraphics[width=\textwidth]{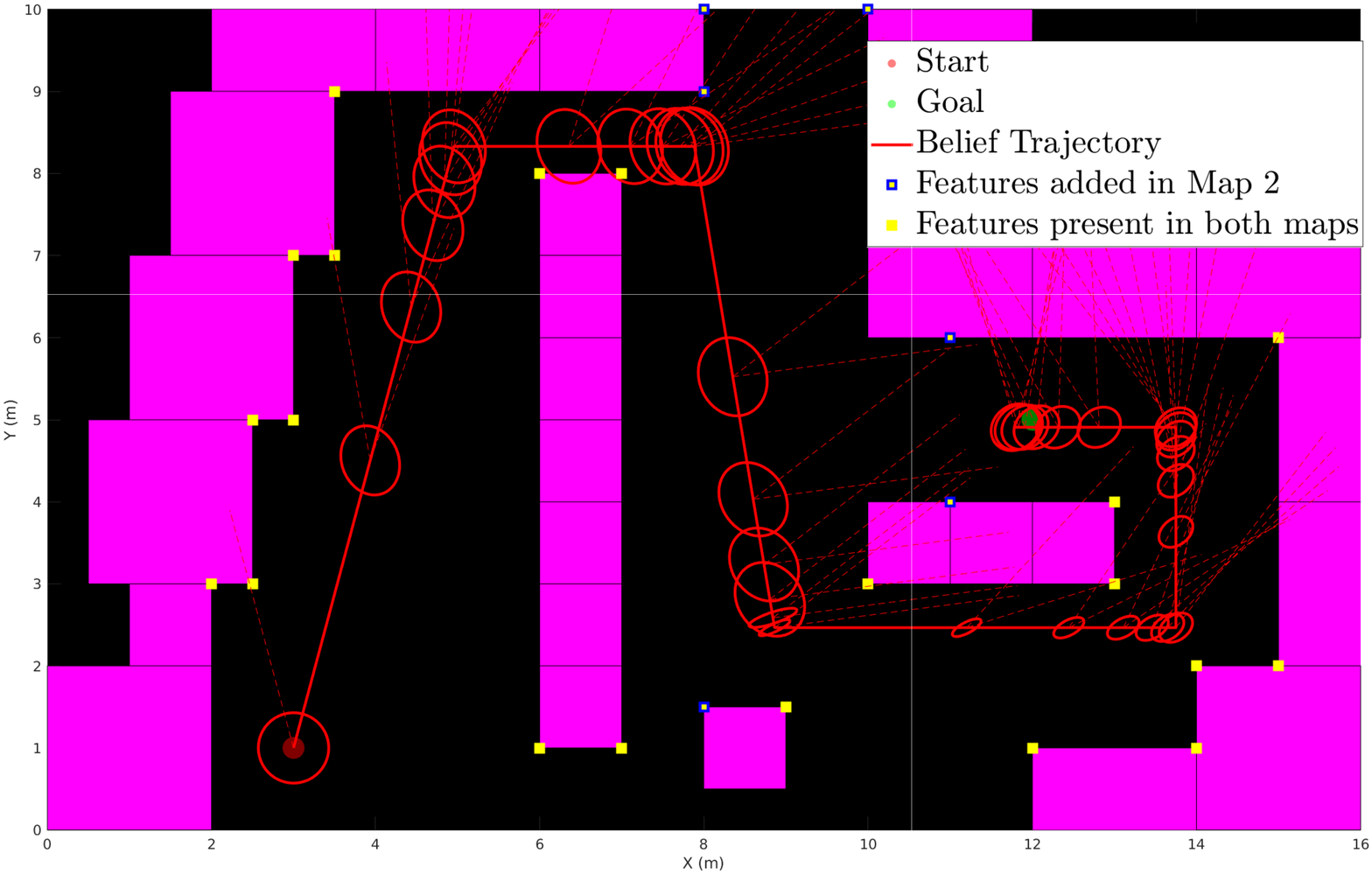}
            \caption{\footnotesize Map 1 and 2 with initial belief trajectory}
            \label{init1}
        \end{subfigure}
        \begin{subfigure}{0.47\textwidth}  
            \includegraphics[width=\textwidth]{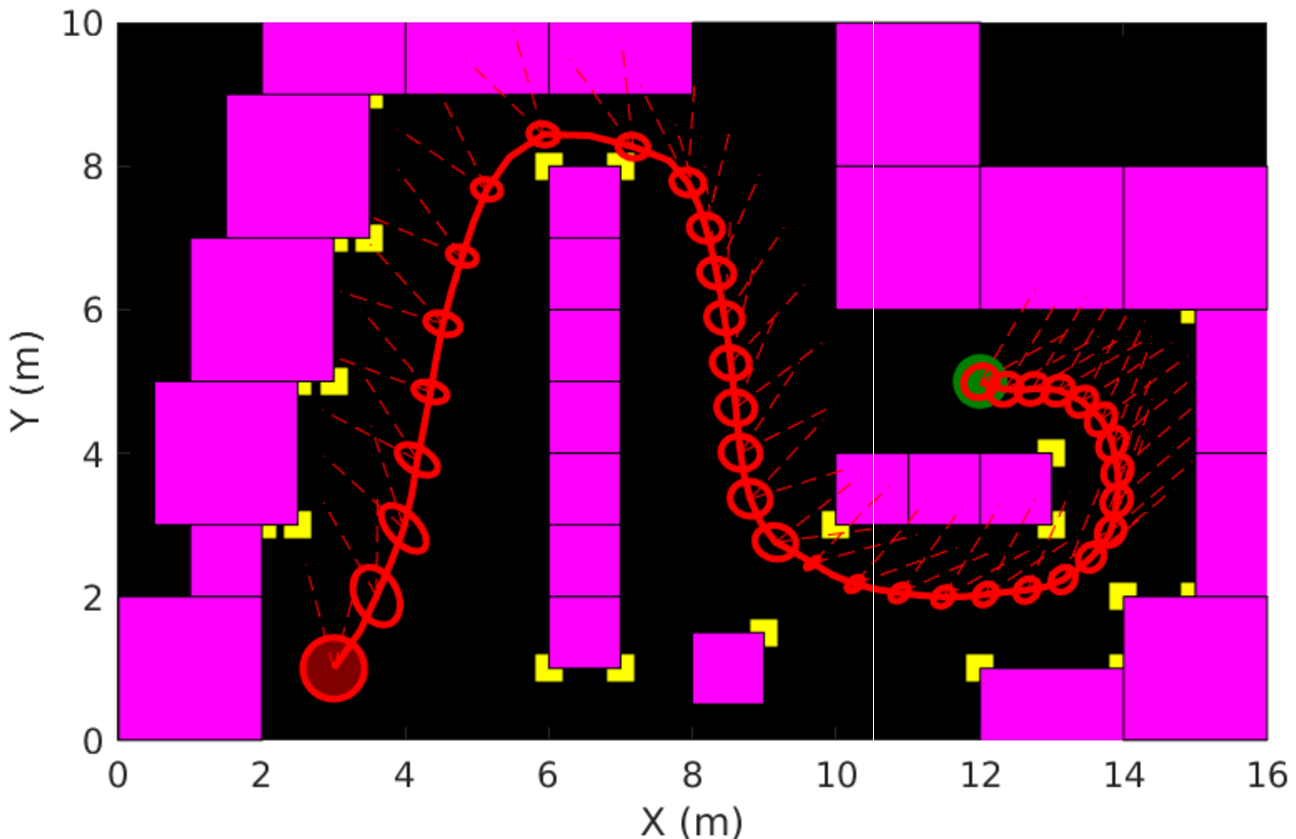}
            \caption{\footnotesize Loose constraints, optimal cost = 4328 units}
            \label{loose}
        \end{subfigure}
        \begin{subfigure}{0.47\linewidth}
            \includegraphics[width=\textwidth]{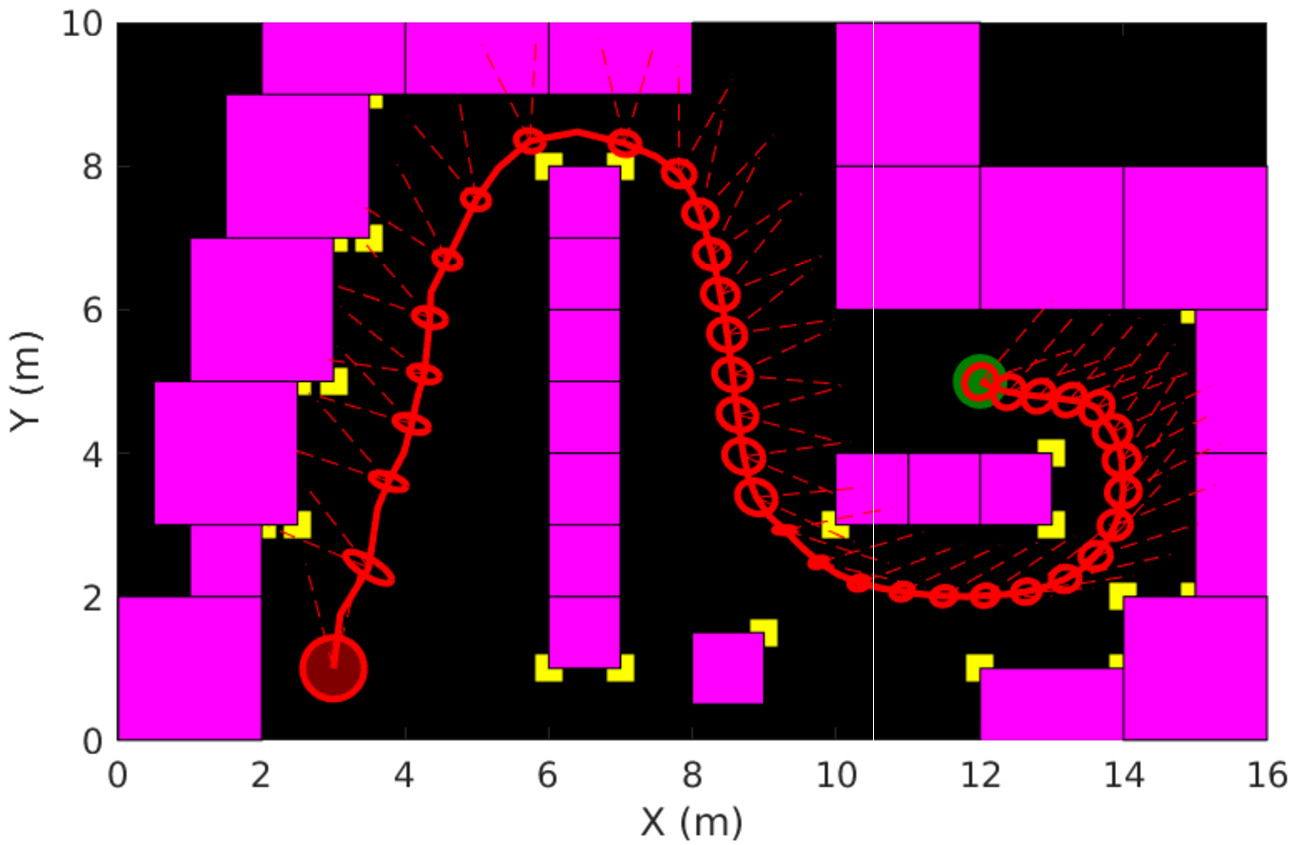}
            \caption{\footnotesize Medium constraints, optimal cost = 4645 units}    
            \label{medium}
        \end{subfigure}
        \begin{subfigure}{0.47\linewidth}  
            \includegraphics[width=\textwidth]{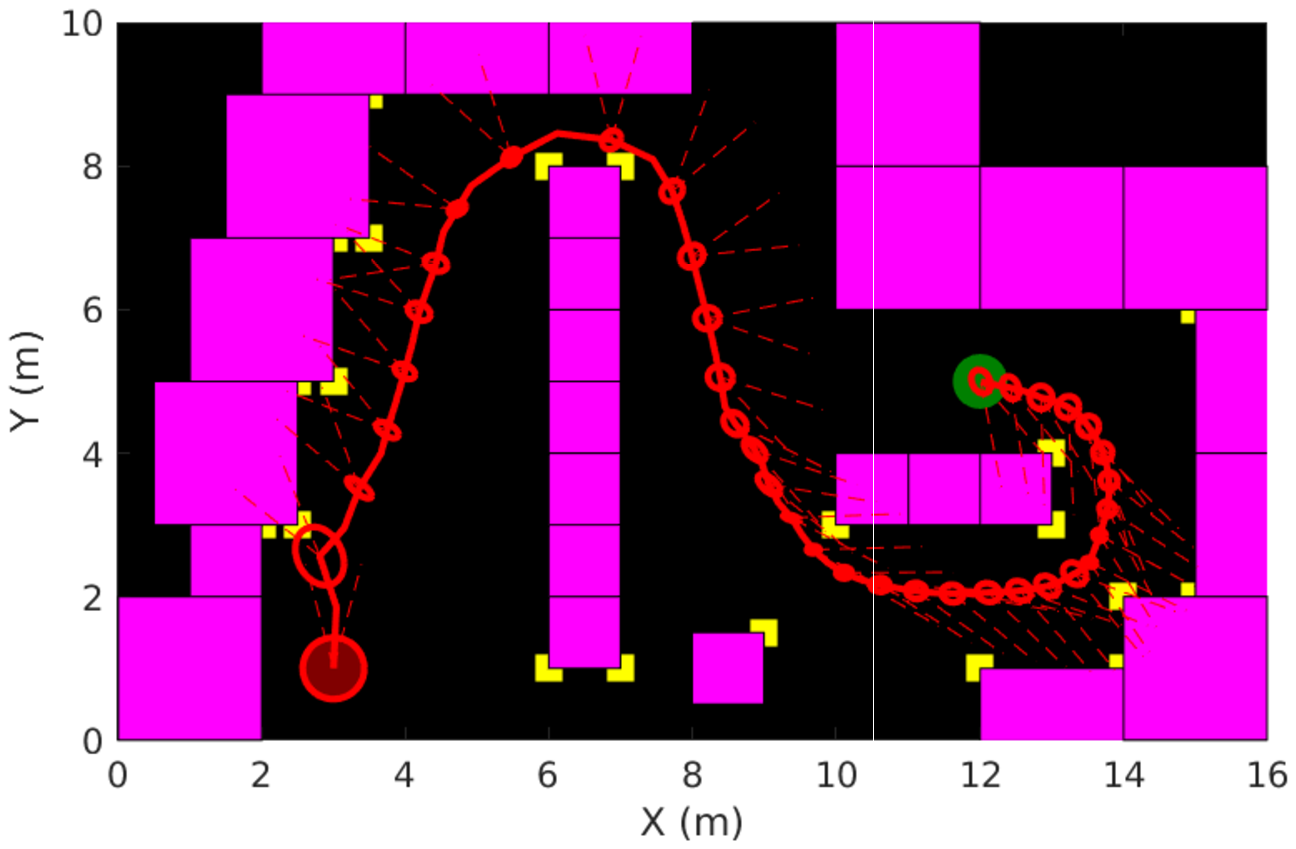}
            \caption{\footnotesize Tight constraints, optimal cost = 7172 units}   
            \label{tight}
        \end{subfigure}
        \caption{iLQG-AL at different constraint levels in Map 1.}
        \label{expt1_traj}
    \end{figure*}
Next, substituting \eqref{eq:v_quad} into \eqref{eq:q_func} and making second order approximations of $\tilde{c}_k(\mbf{b}_k,\mbf{u}_k)$  and first order approximation of the belief dynamics \eqref{eq:bel_dyn} and collecting terms, it can be shown that
\begin{align}\label{eq:q_quad}
    Q_k(\mbf{b}_k,\mbf{u}_k) \approx &Q_k(\bar{\mbf{b}}_k,\bar{\mbf{u}}_k) + Q_{\mbf{b},k}^{T}\delta\mbf{b}_k + Q_{\mbf{u},k}^{T}\delta\mbf{u}_k +\nonumber\\
    &\frac{1}{2}( \delta\mbf{b}_k^{T}Q_{\mbf{bb},k}\delta\mbf{b}_k + 2 \delta\mbf{b}_k^{T}Q_{\mbf{bu},k}\delta\mbf{b}_k \nonumber +\\ &\delta\mbf{u}_k^{T}Q_{\mbf{uu},k}\delta\mbf{u}_k),
\end{align}
with
\begin{align*}
    Q_{\mbf{b},k} &= \tilde{c}_{\mbf{b},k} + \mbf{g}_{\mbf{b},k}^{T}V_{\mbf{b}_{k+1}} + \sum^{p}_{i=1}\mbf{w}^{i^{T}}_{\mbf{b},k}V_{\mbf{bb}_{k+1}}\bar{\mbf{w}}^{i}_{k},\\
    Q_{\mbf{u},k} &= \tilde{c}_{\mbf{u},k} + \mbf{g}_{\mbf{u},k}^{T}V_{\mbf{b}_{k+1}} + \sum^{p}_{i=1}\mbf{w}^{i^{T}}_{\mbf{u},k}V_{\mbf{bb}_{k+1}}\bar{\mbf{w}}^{i}_{k}\\
    Q_{\mbf{bb},k} &= \tilde{c}_{\mbf{bb},k} + \mbf{g}_{\mbf{b},k}^{T}V_{\mbf{bb}_{k+1}}\mbf{g}_{\mbf{b},k} + \sum^{p}_{i=1}\mbf{w}^{i^{T}}_{\mbf{b},k}V_{\mbf{bb}_{k+1}}\mbf{w}^{i}_{\mbf{b},k},\\
    \quad Q_{\mbf{uu},k} &=\tilde{c}_{\mbf{uu},k} + \mbf{g}_{\mbf{u},k}^{T}V_{\mbf{bb}_{k+1}}\mbf{g}_{\mbf{u},k} + \sum^{p}_{i=1}\mbf{w}^{i^{T}}_{\mbf{u},k}V_{\mbf{bb}_{k+1}}\mbf{w}^{i}_{\mbf{u},k}\\
    Q_{\mbf{bu},k} &= \tilde{c}_{\mbf{bu},k} + \mbf{g}_{\mbf{u},k}^{T}V_{\mbf{bb}_{k+1}}\mbf{g}_{\mbf{b},k} + \sum^{p}_{i=1}\mbf{w}^{i^{T}}_{\mbf{u},k}V_{\mbf{bb}_{k+1}}\mbf{w}^{i}_{\mbf{b},k}, \\
    Q_k(\bar{\mbf{b}}_k,\bar{\mbf{u}}_k) &= \tilde{c}_k(\bar{\mbf{b}}_k,\bar{\mbf{u}}_k) + \bar{V}_{k+1} + \frac{1}{2}\sum^{p}_{i=1}\bar{\mbf{w}}^{i^{T}}_{k}V_{\mbf{bb}_{k+1}}\bar{\mbf{w}}^{i}_{k}
\end{align*}
where $\bar{\mbf{w}} = \mbf{w}(\bar{\mbf{b}}_k,\bar{\mbf{u}}_k),p = \mathrm{dim}(\bm{\xi}_k)$ and $(\cdot)^i$ extracts the $i$\textsuperscript{th} column. Thus to find the optimal control update $\delta\mbf{u}^{*}_k$, we take the derivative of \eqref{eq:q_quad} with respect to $\delta\mbf{u}_k$ and set the resulting expression to zero to obtain the affine state feedback law
\begin{equation}\label{eq:unc_law}
    \delta\mbf{u}^{*}_k = -Q_{\mbf{uu},k}^{-1}Q_{\mbf{u},k}  -Q_{\mbf{uu},k}^{-1}Q_{\mbf{bu},k}\mbf{b}_k.
\end{equation}
Thus the full policy  at timestep $k$ is given by
\begin{align}\label{eq:policy_k}
    \mbf{u}_k &= \bm{\pi}_k(\mbf{b}_k) = \bar{\mbf{u}}_k + \delta\mbf{u}^{*}_k\nonumber\\
              &= \bar{\mbf{u}}_k  -Q_{\mbf{uu},k}^{-1}Q_{\mbf{u},k}  -Q_{\mbf{uu},k}^{-1}Q_{\mbf{bu},k}(\mbf{b}_k - \bar{\mbf{b}}_k)
\end{align}
\eqref{eq:unc_law} is then substituted into \eqref{eq:q_quad} and terms collected to obtain
\begin{align}\label{eq:v_update}
  \bar{V}_k &= Q_k(\bar{\mbf{b}}_k,\bar{\mbf{u}}_k) - \frac{1}{2}Q_{\mbf{u},k}^{T}Q_{\mbf{uu},k}^{-1}Q_{\mbf{u},k}\\
  V_{\mbf{b}_k} &= Q_{\mbf{b},k} - Q_{\mbf{bu},k}^{T}Q_{\mbf{uu},k}^{-1}Q_{\mbf{u},k}\\
  V_{\mbf{bb}_k} &= Q_{\mbf{bb},k} - Q_{\mbf{bu},k}^{T}Q_{\mbf{uu},k}^{-1} Q_{\mbf{bu},k}
\end{align}
which are required to compute the quadratic approximation of the value function at time $k$. The above procedure is then recursively repeated back through the time horizon. 

This concludes one iteration of the ``inner" loop. At the beginning of the next iteration, we compute a new nominal belief and control trajectory by applying the new policy \eqref{eq:policy_k} from $\mbf{b}_0$ through the belief dynamics \eqref{eq:bel_dyn} which is known as the ``forward pass" of the iLQG method. The procedure is repeated with all approximations about this new nominal belief and control trajectory. The inner loop terminates once standard convergence conditions are met.  
\subsection{Outer Loop: Updating the penalty function}\label{sec:outer}
Once a solution to the unconstrained problem in \eqref{eq:aug_prob} is computed, the solution is checked for progress towards feasibility against a set of user defined threshold parameters. More specifically, we check if ${\psi}_{j,k}(\mbf{b}_k,\mbf{u}_k) < \phi_{j,k}$, where  ${\psi}_{j,k}$ is the value of the $j$\textsuperscript{th} constraint function and $\phi_{j,k}>0$ is the threshold parameter for the $j$\textsuperscript{th} constraint at time $k$. If the check is successful, the Lagrange multipliers are updated via 
\begin{equation}\label{eq:mult_update}
    \bm{\lambda}_k \leftarrow \frac{\partial}{\partial \bm{\psi}_{k}} \mathcal{P}\left(\bm{\lambda}_{k}, \bm{\mu}_{k}, \bm{\psi}_{k}\left(\mbf{b}_{k}, \mbf{u}_{k}\right)\right),
\end{equation}
which comes from comparing the \textit{Karush–Kuhn–Tucker} (KKT) conditions for optimality for the true Lagrangian of \eqref{big_opt} with the Augmented Lagrangian of \eqref{eq:aug_prob}. The threshold parameters are then reduced while keeping the penalty parameters fixed, fully defining the new unconstrained subproblem to be solved via the method in Section \ref{sec:inner}.
If ${\psi}_{j,k}(\mbf{b}_k,\mbf{u}_k) > \phi_{j,k}$, sufficient constraint improvement has not been achieved. In this case, $\lambda_{j,k}$ is held fixed and $\mu_{j,k}$ is increased, leading to the next subproblem having a greater weight on constraint violation. Although there are many different variations on the updating schedules of $\mu_{j,k}$ and $\phi_{j,k}$ in the literature, most suggest a monotonically decreasing schedule for the $\phi_{j,k}$ and monotonically increasing schedule for $\mu_{j,k}$~\cite{nocedal2006numerical}.
\begin{figure}
    \centering
        \begin{subfigure}{\linewidth}
            \includegraphics[width=0.9\textwidth]{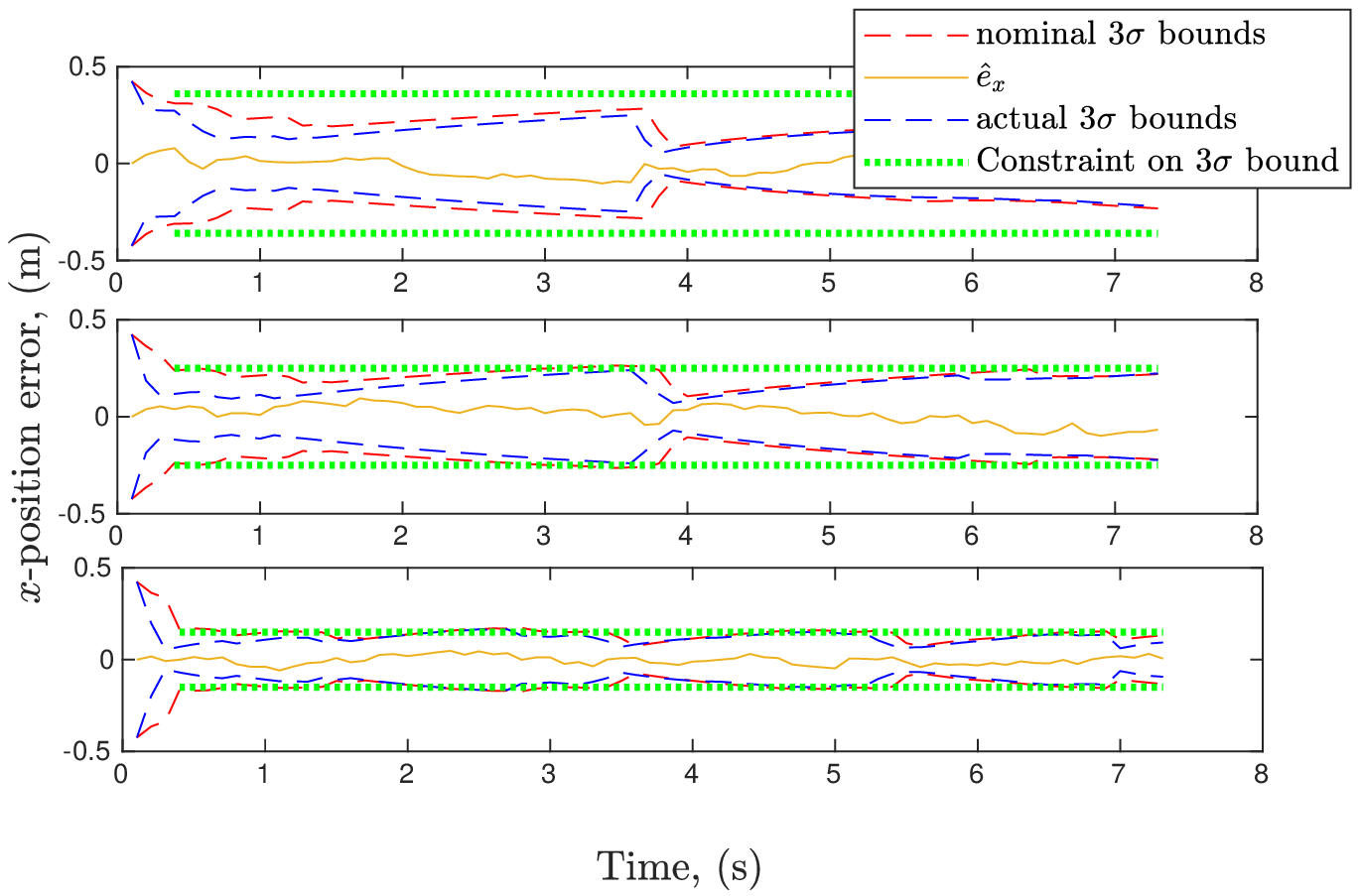}
            \caption{\footnotesize $x$ direction errors}
            \label{loose_err}
        \end{subfigure}
        \begin{subfigure}{\linewidth}  
            \includegraphics[width=0.9\textwidth]{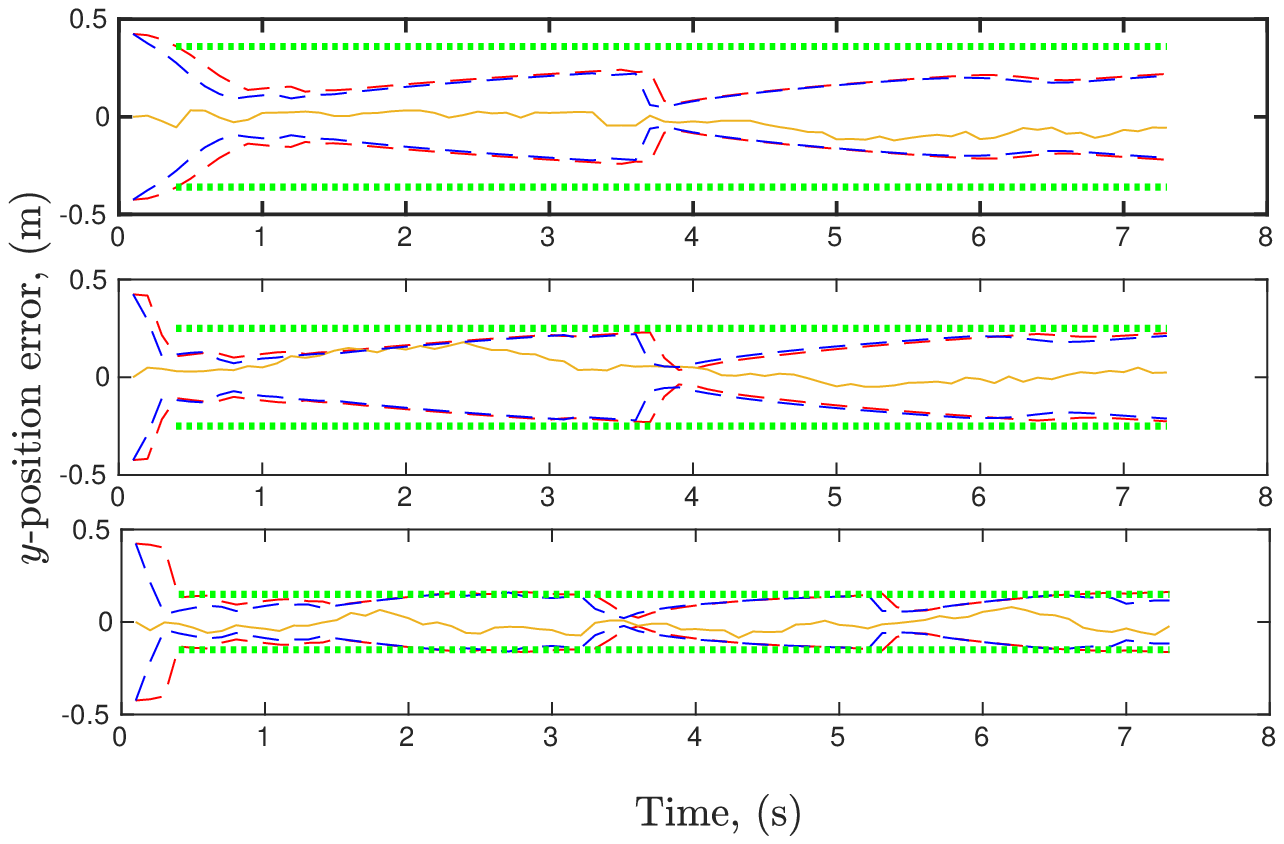}
            \caption{\footnotesize $y$ direction errors}
            \label{medium_err}
        \end{subfigure}
        \begin{subfigure}{\linewidth}
            \includegraphics[width=0.9\textwidth]{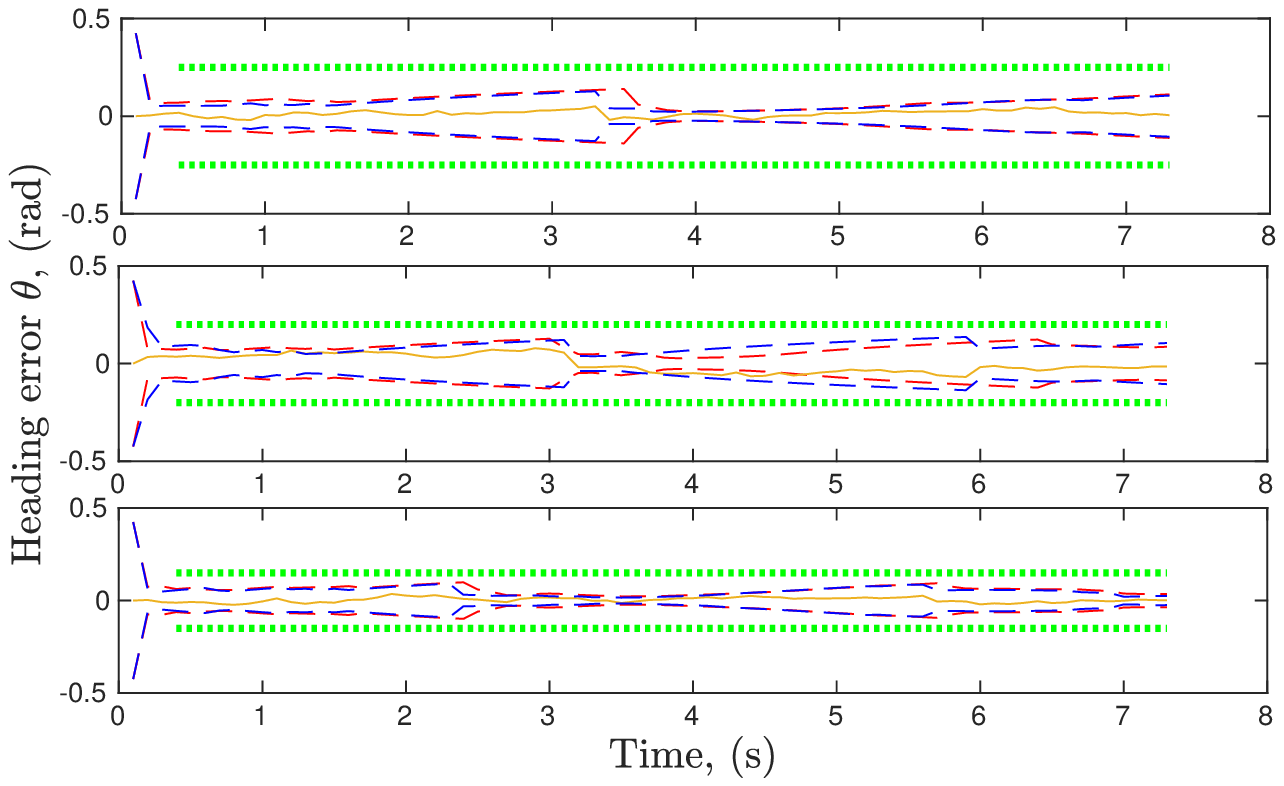}
            \caption{\footnotesize $\theta$ direction errors}    
            \label{tight_err}
        \end{subfigure}
        \caption{Estimation errors at execution time for iLQG-AL at loose, medium and tight level of constraints.}
        \label{fig:errors}
    \end{figure}
\section{Results}\label{sec:exp}
\begin{figure}[th]
    \centering
    \includegraphics[width=0.47\textwidth]{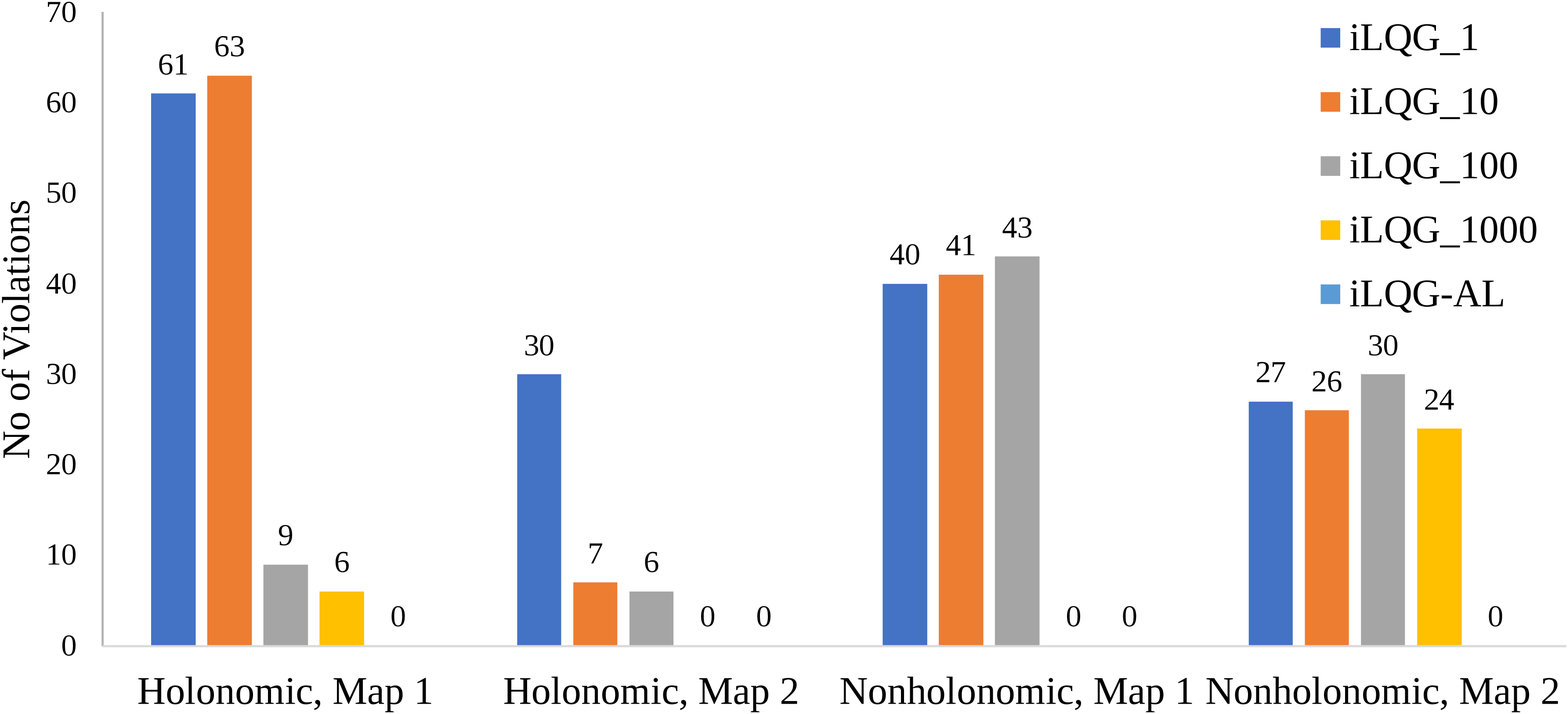}
    \caption{No. of violations for different weights of unconstrained iLQG and iLQG-AL in Maps 1 and 2 and for different robot models.}
    \label{tab:violations}
\end{figure}
Our algorithm was tested in two similar 2-D simulation environments with known obstacle and slightly different feature locations, named Maps 1 and 2 respectively and shown in Figure \ref{expt1_traj}. Two different robot models were tested, a holonomic 2-D disk robot and a nonholonomic  2-D unicycle robot. Both models have the same state, $\mbf{x} = [x,y,\theta]^T$. For the holonomic robot, the inputs $\mbf{u} = [u_x, u_y, \omega]$ and disturbances $\mbf{v} = [v_x, v_y, v_{\omega}]$ are $x$,$y$ velocity and turning rate respectively. For the non-holonomic robot, the inputs $\mbf{u} = [u,\omega]$ and disturbances $\mbf{v} = [v_u, v_{\omega}]$ are forward speed and turning respectively. Their corresponding motion models are
\begin{equation}
    \mbf{x}_{k+1} = \mbf{f}(\mbf{x}_k,\mbf{u}_k,\mbf{v}_k) = \mbf{x}_{k} + \Delta T (\mbf{u}_k + \mbf{v}_k)
\end{equation} 
and
\begin{equation}
    \mbf{x}_{k+1} = \mbf{x}_k + \bma{cc}
    \cos{\theta_k} & 0\\
    \sin{\theta_k} & 0\\
    0 & 1
    \ema \Delta T (\mbf{u}_k + \mbf{v}_k)
\end{equation}
respectively, where $\Delta T$ is the length of the timestep. The measurement consists of a calibrated stereo camera as in Equation~\eqref{eq:cam_meas} with a fixed 30 degree FoV. Since the environment is 2-D, only the horizontal pixel measurements are considered as all features are assumed to be on the same horizontal plane.

For our experiments, we run two methods: unconstrained iLQG in belief space~\cite{van2012motion} and our proposed  iLQG with Augmented Lagrangian (iLQG-AL). For the former, the cost function is
\begin{equation}\label{eq:costs_exp}
\begin{aligned} 
c_{K}\left[\mathbf{b}_{K}\right] &=(\mbf{x}_G - \hat{\mathbf{x}}_{K})^{T} \mbf{S}_{K} (\mbf{x}_G - \hat{\mathbf{x}}_{K})+\operatorname{tr}(\mbf{S}_{I}\bm{\Sigma}_K) \\ c_{k}\left[\mathbf{b}_{k}, \mathbf{u}_{k}\right] &=\mathbf{u}_{k}^{T} \mbf{S}_{u} \mathbf{u}_{k}+\operatorname{tr}(\mbf{S}_{I}\bm{\Sigma}_k) +s_{c} \sum_{i}\exp(-d_i(\mbf{b}_k))
\end{aligned},
\end{equation}
where $\mbf{x}_G$ is the goal state and $\mbf{S}_{K}, \mbf{S}_{I}$ and $\mbf{S}_{u}$ are diagonal weight matrices associated with terminal state cost, information cost and input costs respectively. The final term in the stage cost is a cost on probability of collision, where $d_i(\mbf{b}_k)$ is the distance between the state mean $\hat{\mbf{x}}$ and the $i$\textsuperscript{th} obstacle, divided by the maximum eigenvalue of the covariance matrix of $\mbf{b}_k$. Thus the collision cost penalizes the mean being close to an obstacle, as well as having a large uncertainty near an obstacle and the weight $s_c$ allows this cost to be tuned relative to the others.

For our iLQG-AL method, the costs are the same except that the information costs are removed, since they are handled as hard constraints. In particular, the constraints are
\begin{equation}\label{eq:const1}
    \bm{\psi}_{k}\left(\mbf{b}_{k}, \mbf{u}_{k}\right) = \bm{\psi}_{k}\left(\mbf{b}_{k}\right) = \mbf{A}\mbf{b}_k - \bm{\sigma}_{max}
\end{equation}
where the matrix $\mbf{A}$ picks out the covariance components of the belief to be bounded, and $\bm{\sigma}_{max} = [\sigma^{2}_{x},\sigma^{2}_{y},\sigma^{2}_{\theta}]$ are the actual desired upper bounds on the elements of the covariance matrix, set by the user.

All of the simulations are performed in MATLAB 2019b on a CPU architecture of i7- 8700K (3.7 GHz). All derivatives in \eqref{eq:q_func} are computed via finite differences using Simpson's rule. 
\subsection{iLQG-AL at varying levels of constraints}
\begin{figure*}
    \centering
        \begin{subfigure}{0.47\textwidth}
            \includegraphics[width=\textwidth]{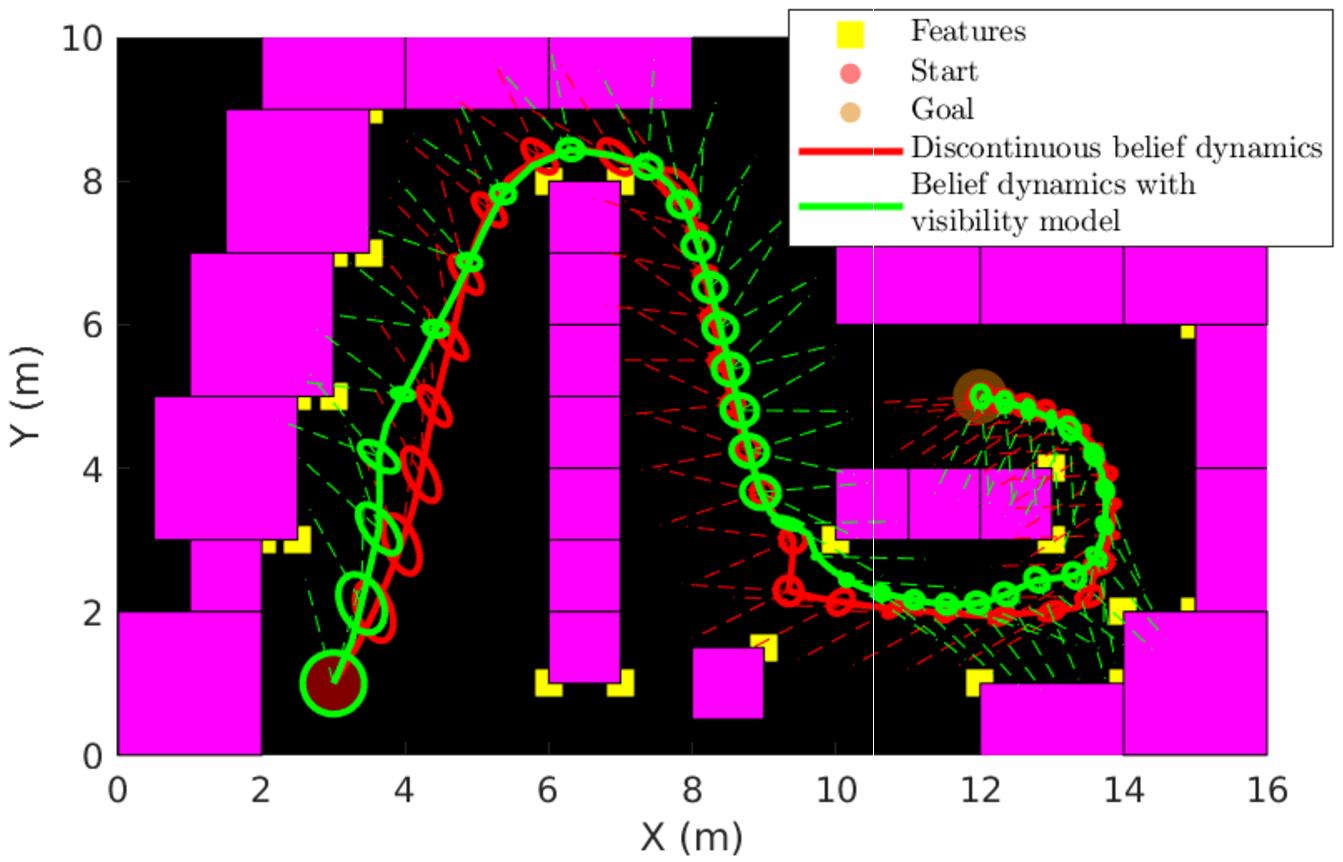}
            \caption{\footnotesize Planned trajectories for the holonomic robot. Optimal cost with and without visibility model: 4647, 5369.  }
            \label{Fig:vis_holo}
        \end{subfigure}
        \begin{subfigure}{0.47\textwidth}  
            \includegraphics[width=\textwidth]{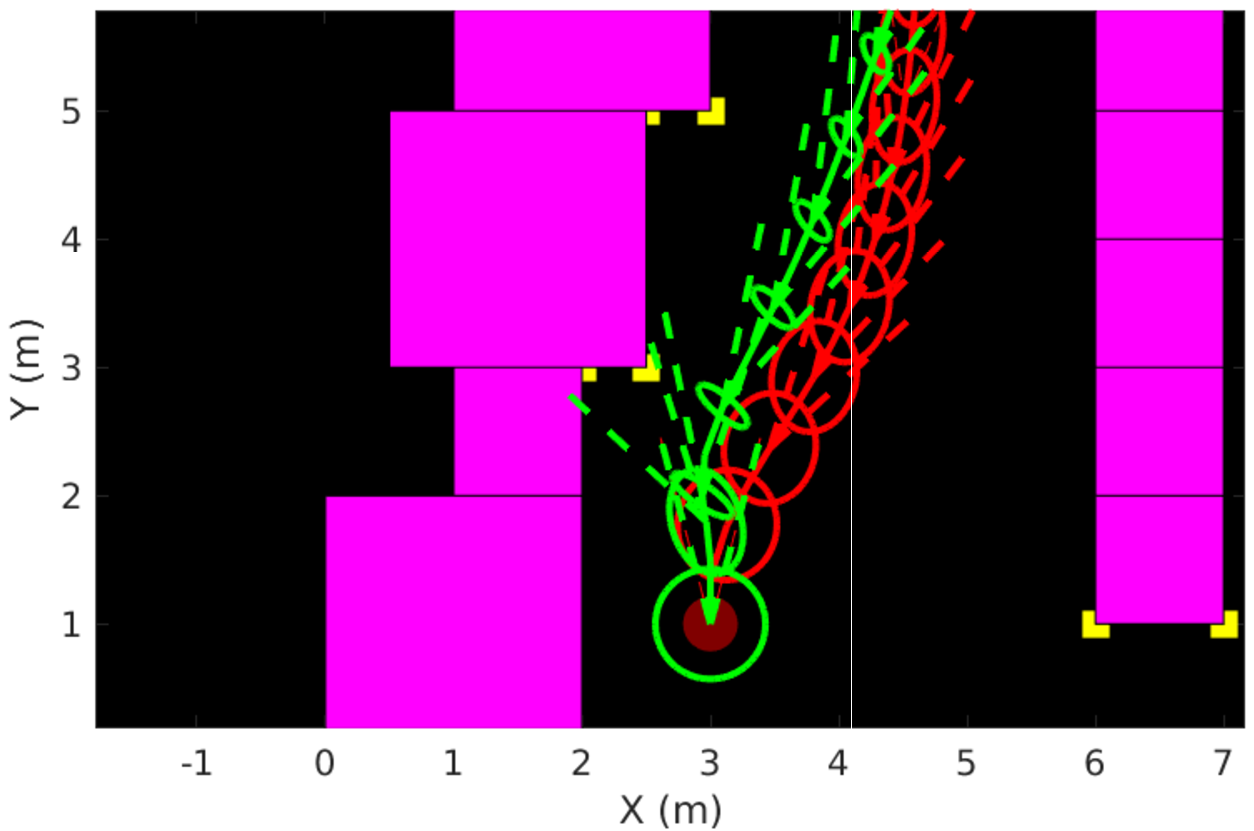}
            \caption{\footnotesize Zoomed in view of the first section of planned trajectories for the nonholonomic robot. Optimal cost with and without visibility model: 4078, 5096.}
            \label{Map2}
        \end{subfigure}
        \caption{Planned belief trajectories with (green) and without (red) visibility modelling.}
        \label{Fig:vis_nonholo}
    \end{figure*}
First, we demonstrate that iLQG-AL can satisfy uncertainty constraints of different levels, and the quantitative and qualitative effects of doing so. We test our algorithm at ``loose", ``medium" and ``tight" constraints, that is, desired upper bounds on $3\sigma_x = 3\sigma_y$ and $3\sigma_{\theta}$ are $[0.36m,0.25 rad],[0.25m,0.2 rad]$ and $[0.15m,0.15 rad]$ respectively. The initial and optimal nominal trajectories for the holonomic robot are shown in Figure \ref{expt1_traj}. Figure \ref{fig:errors} show the optimal covariance at planning time and under measurement noise and disturbances at execution time, as well as the estimation error $\mbf{x}_k - \hat{\mbf{x}}_k$.

As can be seen, the constrained optimization naturally leads to significant changes in camera viewpoints to aggressively focus on nearby features when a tighter uncertainty bound is required, where as looser constraints lead to smoother changes in both path and viewing angle, with no tedious tuning involved. The sharp turns in the trajectories are due to the lack of any path smoothing terms in the cost which can easily be included in practice. The final cost values are also as expected, tighter constraints incur a higher optimal cost in order to be satisfied. 
\subsection{Soft Uncertainty Costs vs Hard Uncertainty Constraints}
Next, we demonstrate the need for dealing with uncertainty limits as inequality constraints as opposed to soft uncertainty costs, that is, the belief space iLQG method of Van der Berg \textit{et al.}, through the following application-inspired scenario. Consider Maps 1 and 2, which have slightly different number and locations of features, shown in Figure \ref{init1}, with the same start and goal points for both the holonomic and nonholonomic robots. The application requires that a maximum estimation uncertainty of  $3\sigma_x = 3\sigma_y = 0.25m$ and $3\sigma_{\theta} = 0.2rad$ is allowed for safe operation. We test both our method, where we simply set the required constraint, vs the unconstrained iLQG method at 4 different weights on uncertainty through the $\mbf{S}_{I}$ matrix in \eqref{eq:costs_exp}. 

The results, summarized in Table \ref{tab:violations}, shows that for the unconstrained iLQG, the same weight on uncertainty costs can result in varying and unpredictable values of uncertainty for different feature locations and robot types. Furthermore, we have only tried one type of scalar cost on a covariance matrix, that is, the trace, whereas the entropy or Rényi entropy~\cite{carrillo2015autonomous} could also be used, which could have different but still unpredictable effects on uncertainty. iLQG-AL can satisfy uncertainty constraints that have intuitive and quantifiable requirements, and the same method can be used in varying environments and hardware without hand tuning.   
\subsection{Effect of probabilistic visibility modelling}
Finally, we investigate the effect of our probabilistic visibility modelling outlined in Section \ref{sec:prob_mod}. We compare the performance of the unconstrained iLQG with and without our visibility model as a part of the belief dynamics, for both the holonomic and nonholonomic robots. The cost function and initial trajectory is the same for both methods.

The costs and resulting trajectories in Figure \ref{Fig:vis_nonholo} show that incorporating visibility into the belief dynamics leads to trajectories that more aggressively change viewpoints, with both lower uncertainty and cost. Furthermore, effects of occlusion are reduced, as seen in the middle section of the trajectory for the holonomic robot (Figure \ref{Fig:vis_holo}), where, with the discontinuous dynamics, the planner fails to change viewpoint even though the features are hidden behind obstacles. As described in Section \ref{sec:prob_mod}, we believe this is due to numerical gradient approximations over discontinuous cost boundaries leading to poor optimization performance, which explains the lower cost achieved with our smooth dynamics.
\section{Conclusion and Further Work}
In this paper, we have presented a novel formulation of the planning under uncertainty problem as an uncertainty constrained trajectory optimization problem in belief space. Furthermore we adopt an Augmented Lagrangian based iLQG method in belief space that can solve the problem, as well as a probabilistic visibility model that helps account for field of view and feature matching limits. 

Future work will involve combining the local trajectory optimization methods with global sampling based methods to address the fact that our method is confined to local minima. We also wish to formulate collision avoidance and input limits as inequality constraints instead of costs. Finally, we wish to test our method on a real robot such as a quadrotor and scale up to realistic map sizes while maintaining realtime performance.


\addtolength{\textheight}{-12cm}   



\bibliographystyle{ieeetr}
\bibliography{IEEEabrv.bib}

\begin{thebibliography}{10}

\bibitem{van2012motion}
J.~Van Den~Berg, S.~Patil, and R.~Alterovitz, ``Motion planning under
  uncertainty using iterative local optimization in belief space,'' {\em The
  International Journal of Robotics Research}, vol.~31, no.~11, pp.~1263--1278,
  2012.

\bibitem{hauskrecht2000value}
M.~Hauskrecht, ``Value-function approximations for partially observable markov
  decision processes,'' {\em Journal of Artificial Intelligence Research},
  vol.~13, pp.~33--94, 2000.

\bibitem{pineau2003point}
J.~Pineau, G.~Gordon, S.~Thrun, {\em et~al.}, ``Point-based value iteration: An
  anytime algorithm for pomdps,'' in {\em International Joint Conference on
  Artificial Intelligence}, vol.~3, pp.~1025--1032, 2003.

\bibitem{chaudhari2013sampling}
P.~Chaudhari, S.~Karaman, D.~Hsu, and E.~Frazzoli, ``Sampling-based algorithms
  for continuous-time pomdps,'' in {\em American Control Conference},
  pp.~4604--4610, 2013.

\bibitem{somani2013despot}
A.~Somani, N.~Ye, D.~Hsu, and W.~S. Lee, ``Despot: Online pomdp planning with
  regularization,'' in {\em Advances in Neural Information Processing Systems},
  pp.~1772--1780, 2013.

\bibitem{shani2013survey}
G.~Shani, J.~Pineau, and R.~Kaplow, ``A survey of point-based pomdp solvers,''
  {\em Autonomous Agents and Multi-Agent Systems}, vol.~27, no.~1, pp.~1--51,
  2013.

\bibitem{prentice2009belief}
S.~Prentice and N.~Roy, ``The belief roadmap: Efficient planning in belief
  space by factoring the covariance,'' {\em The International Journal of
  Robotics Research}, vol.~28, no.~11-12, pp.~1448--1465, 2009.

\bibitem{bry2011rapidly}
A.~Bry and N.~Roy, ``Rapidly-exploring random belief trees for motion planning
  under uncertainty,'' in {\em IEEE International Conference on Robotics and
  Automation}, pp.~723--730, 2011.

\bibitem{agha2011firm}
A.-A. Agha-Mohammadi, S.~Chakravorty, and N.~M. Amato, ``Firm: Feedback
  controller-based information-state roadmap-a framework for motion planning
  under uncertainty,'' in {\em IEEE/RSJ International Conference on Intelligent
  Robots and Systems}, pp.~4284--4291, 2011.

\bibitem{agha2015simultaneous}
A.-a. Agha-mohammadi, S.~Agarwal, S.~Chakravorty, and N.~M. Amato,
  ``Simultaneous localization and planning for physical mobile robots via
  enabling dynamic replanning in belief space,'' {\em arXiv preprint
  arXiv:1510.07380}, 2015.

\bibitem{platt2010belief}
R.~Platt~Jr, R.~Tedrake, L.~Kaelbling, and T.~Lozano-Perez, ``Belief space
  planning assuming maximum likelihood observations,'' 2010.

\bibitem{erez2012scalable}
T.~Erez and W.~D. Smart, ``A scalable method for solving high-dimensional
  continuous pomdps using local approximation,'' {\em arXiv preprint
  arXiv:1203.3477}, 2012.

\bibitem{indelman2015planning}
V.~Indelman, L.~Carlone, and F.~Dellaert, ``Planning in the continuous domain:
  A generalized belief space approach for autonomous navigation in unknown
  environments,'' {\em The International Journal of Robotics Research},
  vol.~34, no.~7, pp.~849--882, 2015.

\bibitem{chaves2015risk}
S.~M. Chaves, J.~M. Walls, E.~Galceran, and R.~M. Eustice, ``Risk aversion in
  belief-space planning under measurement acquisition uncertainty,'' in {\em
  IEEE/RSJ International Conference on Intelligent Robots and Systems},
  pp.~2079--2086, 2015.

\bibitem{magnus2019matrix}
J.~R. Magnus and H.~Neudecker, {\em Matrix differential calculus with
  applications in statistics and econometrics}.
\newblock John Wiley \& Sons, 2019.

\bibitem{minka1998expectation}
T.~Minka, ``Expectation-maximization as lower bound maximization,'' {\em
  Tutorial published on the web at http://www-white. media. mit.
  edu/tpminka/papers/em. html}, vol.~7, p.~2, 1998.

\bibitem{lantoine2012hybrid}
G.~Lantoine and R.~P. Russell, ``A hybrid differential dynamic programming
  algorithm for constrained optimal control problems. part 1: Theory,'' {\em
  Journal of Optimization Theory and Applications}, vol.~154, no.~2,
  pp.~382--417, 2012.

\bibitem{aoyama2020constrained}
Y.~Aoyama, G.~Boutselis, A.~Patel, and E.~A. Theodorou, ``Constrained
  differential dynamic programming revisited,'' {\em arXiv preprint
  arXiv:2005.00985}, 2020.

\bibitem{todorov2005generalized}
E.~Todorov and W.~Li, ``A generalized iterative lqg method for locally-optimal
  feedback control of constrained nonlinear stochastic systems,'' in {\em
  Proceedings of the 2005, American Control Conference, 2005.}, pp.~300--306,
  IEEE, 2005.

\bibitem{nocedal2006numerical}
J.~Nocedal and S.~Wright, {\em Numerical optimization}.
\newblock Springer Science \& Business Media, 2006.

\bibitem{carrillo2015autonomous}
H.~Carrillo, P.~Dames, V.~Kumar, and J.~A. Castellanos, ``Autonomous robotic
  exploration using occupancy grid maps and graph slam based on shannon and
  r{\'e}nyi entropy,'' in {\em 2015 IEEE international conference on robotics
  and automation (ICRA)}, pp.~487--494, IEEE, 2015.

\end{thebibliography}

\end{document}